\documentclass[10pt, a4paper]{article}
\usepackage{lrec}
\usepackage{multibib}
\newcites{languageresource}{Language Resources}
\usepackage{graphicx}
\usepackage{tabularx}
\usepackage{soul}

\usepackage{epstopdf}
\usepackage[utf8]{inputenc}
\usepackage[T1]{fontenc}

\usepackage{hyperref}
\usepackage{xstring}
\usepackage{enumitem}
\usepackage{color}

\title{NegBERT: A Transfer Learning Approach for Negation Detection and Scope Resolution}

\name{Aditya Khandelwal, Suraj Sawant}

\address{College of Engineering Pune \\
         khandelwalar16.comp@coep.ac.in, sts.comp@coep.ac.in \\
         }

\abstract{
Negation is an important characteristic of language, and a major component of information extraction from text. This subtask is of considerable importance to the biomedical domain. Over the years, multiple approaches have been explored to address this problem: Rule-based systems, Machine Learning classifiers, Conditional Random Field models, CNNs and more recently BiLSTMs. In this paper, we look at applying Transfer Learning to this problem. First, we extensively review previous literature addressing Negation Detection and Scope Resolution across the 3 datasets that have gained popularity over the years: the BioScope Corpus, the Sherlock dataset, and the SFU Review Corpus. We then explore the decision choices involved with using BERT, a popular transfer learning model, for this task, and report state-of-the-art results for scope resolution across all 3 datasets. Our model, referred to as NegBERT, achieves a token level F1 score on scope resolution of 92.36 on the Sherlock dataset, 95.68 on the BioScope Abstracts subcorpus, 91.24 on the BioScope Full Papers subcorpus, 90.95 on the SFU Review Corpus, outperforming the previous state-of-the-art systems by a significant margin. We also analyze the model’s generalizability to datasets on which it is not trained. \\ \newline  }

\begin{document}

\maketitleabstract

\section{Introduction}

Negation Detection and Scope Resolution is an important subtask for tasks ranging from Sentiment Analysis, where the sentiment of a given sentence is dependent on negation, to query response systems like Chatbots, where negation entirely changes the meaning and hence the relevance of a certain body of text. A substantial portion of the research till date on this topic focused solely on data from the biomedical domain, where use of negation cues is abundant (eg. medical reports). While negation is intuitive for humans to spot, finding the exact words that indicate such negation and delineating the scope of such negation cues has proven to be a tricky problem for computer-based systems. One could imagine that finding negation cues and their scopes could be easily solved via rules and carefully designed heuristics, and this was the exact approach used by the initial few systems attempting this task. But given the complexities of human language, these approaches weren’t accurate enough. Thus, other methods were explored, and Deep Learning-based approaches have shown to be particularly promising.
\begin{center}
A simple example of negation is as follows:\\
This is \underline{not} [a negation].
\end{center} 
\par We can observe that ‘not’ is the negation word (known as the negation cue) and the words whose meaning is altered by ‘not’ are ‘a’ and ‘negation’, which belong to what is known as the cue’s scope. Negation detection involves finding these negation cues, and scope resolution for each cue necessitates finding the words affected negatively by that cue (finding its scope).
\par Cues can come in a variety of ways:
\begin{enumerate}[noitemsep]
    \item An affix: (im)perfect, (a)typical, ca(n’t)
    \item A single word: not, no, failed, lacks
    \item A set of consecutive words or discontinuous words:
	neither…nor
\end{enumerate}
The scope of a cue is also not constrained to be a continuous sequence of words. These facts, coupled with the relatively small dataset sizes compared to other NLP datasets, make this task particularly challenging to solve.
\par Transfer Learning, a method in which we train Deep Learning systems on huge corpora and then ‘transfer’ or finetune these pretrained architectures on downstream tasks which have a dearth of data, has taken the NLP community by storm, achieving state-of-the-art results on almost every NLP task rhis approach has been applied to. This method was originally used in Computer Vision, by training models on the ImageNet dataset which allowed them to capture important features to look for in a picture, and then apply to other datasets by changing the final layer and training on the downstream task. Recently, a number of architectures including BERT \cite{bert} have applied this to NLP, contributing massively to the advancement of research in the field. Almost every NLP task benefitted from transfer learning, as training on massive corpora allowed these models to learn an understanding of language.
\par Motivated by the success of transfer learning, we apply BERT to negation detection and scope resolution. We explore the set of design choices involved, and experiment on all 3 public datasets available: the BioScope Corpus (Abstracts and Full Papers subcorpora) \cite{szarvas-etal-2008-bioscope}, the Sherlock Dataset \cite{morante-blanco-2012-sem} and the SFU Review Corpus \cite{konstantinova-etal-2012-review}. The Sherlock Corpus was used in the *sem 2012 Shared Task on negation detection and scope resolution.  We train NegBERT on one dataset and report the scores for testing on all datasets, thus showing the generalizability of NegBERT. Since the BioScope dataset is primarily from the biomedical domain, while the Sherlock dataset is taken from stories by Sir Author Conan Doyle (literary work), and the SFU Review Corpus is a collection of product reviews (free text by human users), the 3 datasets belong to different domains.
\par This paper is organized as follows: In Section 2, we extensively review available literature on the subject. Section 3 contains the details of the methodology used for NegBERT, while Section 4 includes experimental details for our experimentation. In Section 5, we report the results of our experimentation and in Section 6, analyze them. In Section 7, we perform an analysis of the errors made by the system with regards to a problem in annotation schema brought out by \newcite{fancellu-etal-2017-detecting}. Our conclusions and our perspective on the future scope for this problem is presented in Section 8.

\section{Previous Work}

In this section, we look at the previous literature addressing this task. The results of the approaches are neatly summarized at the end of the section in Tables \ref{fig:litreviewfig1} and \ref{fig:litreviewfig2}.

\subsection{Rule-Based Approaches}
The first approach that was explored in literature was a simple rule-based system \cite{mutalik}. They tested the hypothesis that a lexical scanner that uses regular expressions to generate a finite state machine can detect negation cues in natural language. Their algorithm, NegFinder, was based on a manual inspection of 40 medical documents. They showed that it was possible to apply computational methods to detect negation cues in a sentence.
\par \newcite{chapman} proposed a simple regular expression algorithm (NegEx) to detect negation cues. They posited that medical language is lexically less ambiguous and hence a rule-based system can be applied, and that a simpler system than the one proposed by \newcite{mutalik} also performed well. NegEx is a very reliable algorithm in the medical domain, which has been extensively used in further research.
\par \newcite{sanchez} looked at negation cue detection in the domain of Protein-Protein Interaction, proposing a heuristic-based system using a full dependency parser to extract negations tailored to that specific domain. They used the fact that rule-based systems had to be domain specific to perform well.
\par \newcite{Huang2007ResearchPA} stated that previous research had shown that the scope of negation may be difficult to identify if the cues are more than a few words away, and hence focused on addressing this problem. They proposed combining regular expression matching with grammatical parsing, which allowed the rule-based systems to account for long-term dependencies.
\par For the *sem 2012 Shared task, the team from UCM-1 \cite{carrillo-de-albornoz-etal-2012-ucm} used a rule-based system to detect negation cues. Scopes were resolved using the syntax tree of the sentence in which the negation arises. Their system was initially intended for processing negation in opinionated texts and was adapted to fit the task requirements. The team from UCM-2 \cite{ballesteros-etal-2012-ucm} relied on a rule-based system engineered to the given dataset. Cue detection was performed via a static cue lexicon, scope was detected using rules based on a prior work by \newcite{ballesteros-etal-2012-ucm} for the BioScope Corpus, which was modified for the Sherlock Corpus. The team from UGroningen \cite{basile-etal-2012-ugroningen} also used a rule-based system based on NLP toolchain used to construct the Groningen Meaning Bank. Their system transformed the texts into logical formulas – using the C\&C tools and Boxer, another system. They concluded that it is not easy to transfer the information about negation from a formal, logical representation of scope to a theory-neutral surface-oriented approach.
\par These 3 methods showed how even rule-based systems with well-defined task-specific rules showed acceptable performance. The primary limitation, of course, were that these rules were not generalizable across domains or even datasets.
\par \newcite{sohn} looked to improve Mayo Clinic’s clinical Text Analysis and Knowledge Extraction System (cTAKES) negation annotator via dependency parsers and a rule-based system. They found that using dependency-based negation proved to be a superior alternative to the pre-existing cTAKES negation annotator.
\par \newcite{MEHRABI2015213} proposed DEEPEN as an improvement to NegEx which added dependency parsing to it. They looked at negation detection only and evaluated their system on the Mayo Clinical Dataset They made NegEx, a very reliable system, much more accurate.
\par More recently, NegBio was introduced \cite{peng}, which utilized Universal Dependency patterns for cue detection. They improved on NegEx and showed that for the medical domain, this performed extremely well. Thus, we observe that recent rule-based systems incorporate dependency parsing in their rules, showing how each word needs to be considered in the context of the words around it. The sequential order of words makes a big difference even in detecting negation.
\subsection{Machine Learning Approaches}

The use of Machine Learning techniques for negation detection was explored by \newcite{rokach} who described an approach to automatically generate and heuristically evaluate Regular Expression patterns. They then fed the results of the pattern evaluations and a few other concept features to a decision tree classifier. They also looked at a cascade of classifiers to make decisions. The sentence was made to pass to the next level of the cascade if no negation was found. The cascade they proposed was 3 levels deep. They relied on the regular expression matching paradigm to generate features but allowed the ML model to use them to come up with better rules(decisions), thus improving on just rule-based systems.
\par In 2008, \newcite{morante-etal-2008-learning} proposed a system to both detect negation and find its scope in biomedical texts. This paper focused on the scope detection task, which hadn’t been previously explored. They proposed a memory-based scope finder that works in 2 phases, cue detection and scope resolution. They used a k-Nearest Neighbors Classifier with features extracted from the sentence and modified to the task at hand. This was a novel approach to negation detection at the time and was performed on the BioScope Corpus.
\par In 2009, \newcite{morante-daelemans-2009-metalearning} used IGTREE, which is a memory-based learning algorithm, as implemented in TiMBL, to detect cues. For scope resolution, they used a metalearner that used the predictions by 3 classifiers which predicted whether a given token was the beginning of a scope, end of the scope of neither. They used a memory-based algorithm, SVM and CRF as the 3 classifiers. This was also done on the Bioscope Corpus and achieved the state-of-the-art results in cue detection on the Bioscope Corpus. This algorithm was majorly rule based for detecting cues, as the algorithm only ran for words that were not a part of a predefined lexicon of words.
\par For the *sem 2012 Shared Task, the team from UABCoRAL: \cite{gyawali-solorio-2012-uabcoral} found the cue using a lexicon and classified each word as in-scope or out-of-scope by extracting features from a 2-tuple of words (the negation cue and the word under consideration) and passing that through a classifier. The team from UiO1: \cite{read-etal-2012-uio1} detected cues in a similar way to \newcite{lapponi-etal-2012-uio}. They used an SVM as the classifier. For scope resolution, they looked at the syntactic units and developed heuristics to improve the system and incorporated a data-driven approach which involved a ranking approach over syntactic constituents. At the time, this outperformed all other algorithms, majority of which were rule-based for the *sem 2012 Shared Task.
\par In 2014, \newcite{packard-etal-2014-simple} looked at negation scope resolution as a semantic problem, and their approach worked over explicit and formal representations of propositional semantics. They proposed an MRS Crawler, and a maximum entropy model for parse ranking, trained on a different dataset of encyclopedia articles and tourist brochures. They achieved the maximum F1 score and outperformed all systems from *sem 2012 Shared Task on the Sherlock dataset. \\
\par In 2015, \newcite{cruznoa} looked at the Simon Fraser University (SFU) Review Corpus. They classified words as per the BIO representation schema. Another classifier attempted to tell if tokens in a sentence are in the scope of a negation cue. They used an SVM classifier with an RBF kernel and used Cost Sensitive Learning to deal with the imbalanced classification.
\par \newcite{oupatrick} looked at negation cue detection and experimented with 3 methods: lexicon-based, syntax-based (both rule-based) and an SVM classifier. The SVM classifier delivered the best results. They collected their own dataset which had data from the biomedical domain. This showed that the most promise was not in furthering rule-based systems, but in exploring ML techniques for negation detection.
\subsection{Conditional Random Field Approaches}
A third approach to this task used the inherent sequential order to a sentence, by using Conditional Random Fields (CRFs). \newcite{agarwalyu} used this approach for scope detection. Their system was robust and could identify scope in both biomedical and clinical domains. \newcite{morante-daelemans-2009-metalearning} had in contrast looked at the task as classification of word pairs, the negation word and the word to be labelled.
\par \newcite{councill-etal-2010-great} looked at negation in the context of improved sentiment analysis. They detected cues using a lexicon of explicit negation cues, and scopes using a CRF model as an annotator in a larger system. While they evaluated their system on the BioScope Corpus, they constructed a corpus called Product Reviews for their task. They showed that training on the biomedical domain and testing on the Product Reviews or vice versa led to poor results. This suggested that the corpora constructed were too small and thus approaches too task-specific to be generalized to natural language.
\par For the *sem 2012 Shared task, the team from UMichigan \cite{abu-jbara-radev-2012-umichigan} trained a CRF on the lexical, structural and syntactic features of the data for both cue detection and scope resolution. They expanded the set of features given to the CRF. The team from FBK \cite{chowdhury-2012-fbk} trained CRF classifiers, trained on only features provided by the dataset. A different set of features was considered for the CRF which exploited phrasal and contextual clues along with token specific features. The team from UiO2: \cite{lapponi-etal-2012-uio} detected cues by maintaining a corpus and classifying known cue words as cue or non-cue. Scope was detected using CRFs trained on lexical and syntactic features, together with a fine-grained set of labels that captured the scopal behavior of certain tokens. The team from UWashington \cite{white-2012-uwashington} detected cues using regular expression rules from the training data. Scope tokens were detected using a CRF sequence tagger and custom defined features fed to the CRF.
\par \newcite{li-lu-2018-learning} used models based on CRFs, semi-Markov CRF and latent-variable CRF, and achieved better results than previously reported on the Sherlock dataset, beating out all deep learning-based systems as well. Their key observation was that certain useful information such as features related to negation cue, long distance dependencies as well as some latent structural information could be exploited for such a task.
\subsection{Reinforcement Learning Approaches}
A distinctive and unique approach to negation scope resolution was the application of reinforcement learning. \newcite{PROLLOCHS201667} looked at negation detection in the context of a decision support system for sentiment analysis. Their system thus represented the state by the encoding of the position in a sentence, and the set of actions as setting the state to negated or not negated. Thus, each token was labelled by the system by taking an action given the current state. This approach did not work as well as one would have hoped.

\begin{table}
    \centering
    \includegraphics[width=0.9\linewidth]{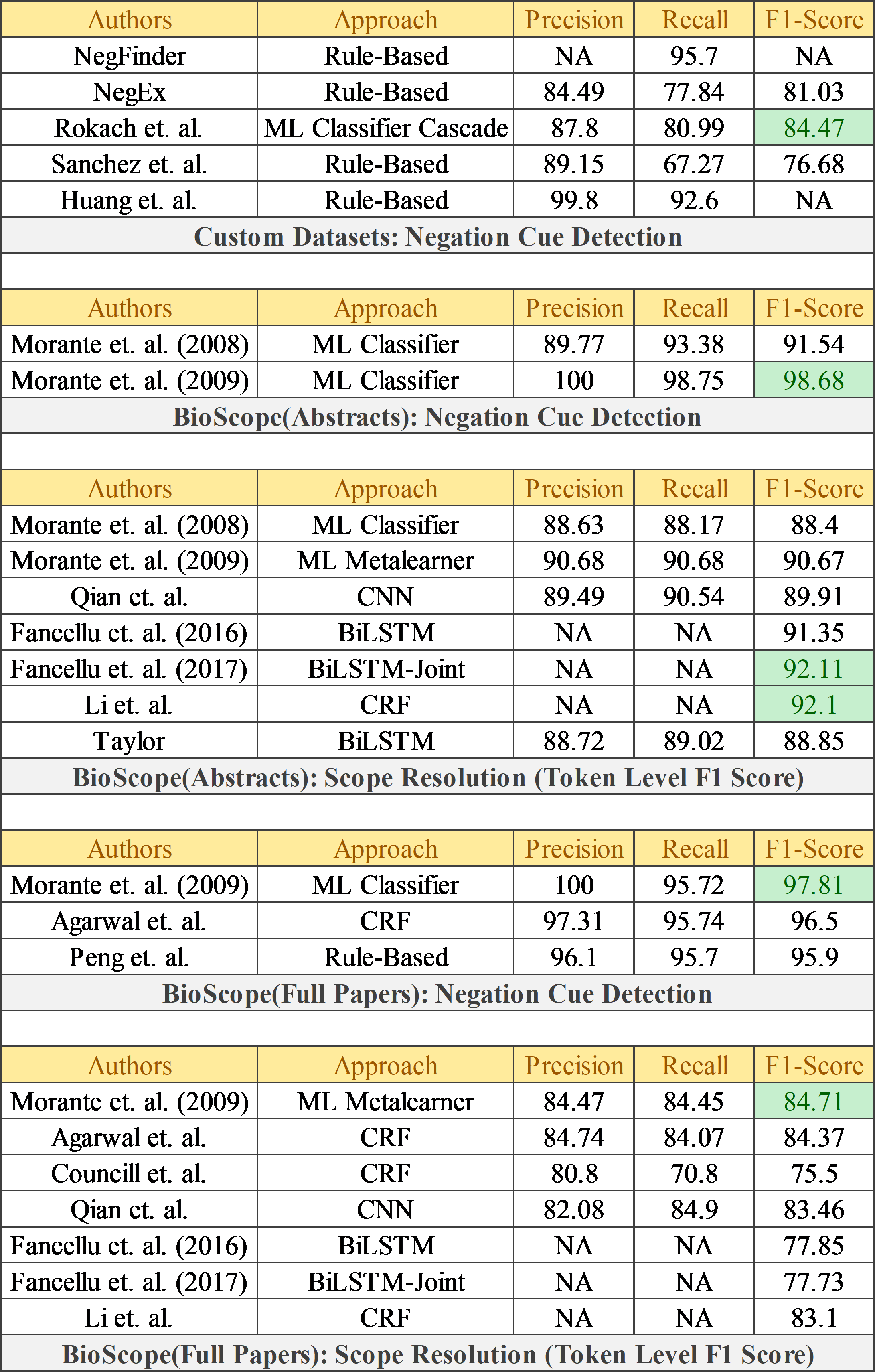}
    \caption{Literature Review: Results Summary}
    \label{fig:litreviewfig1}
\end{table}

\begin{table}
    \centering
    \includegraphics[width=0.9\linewidth]{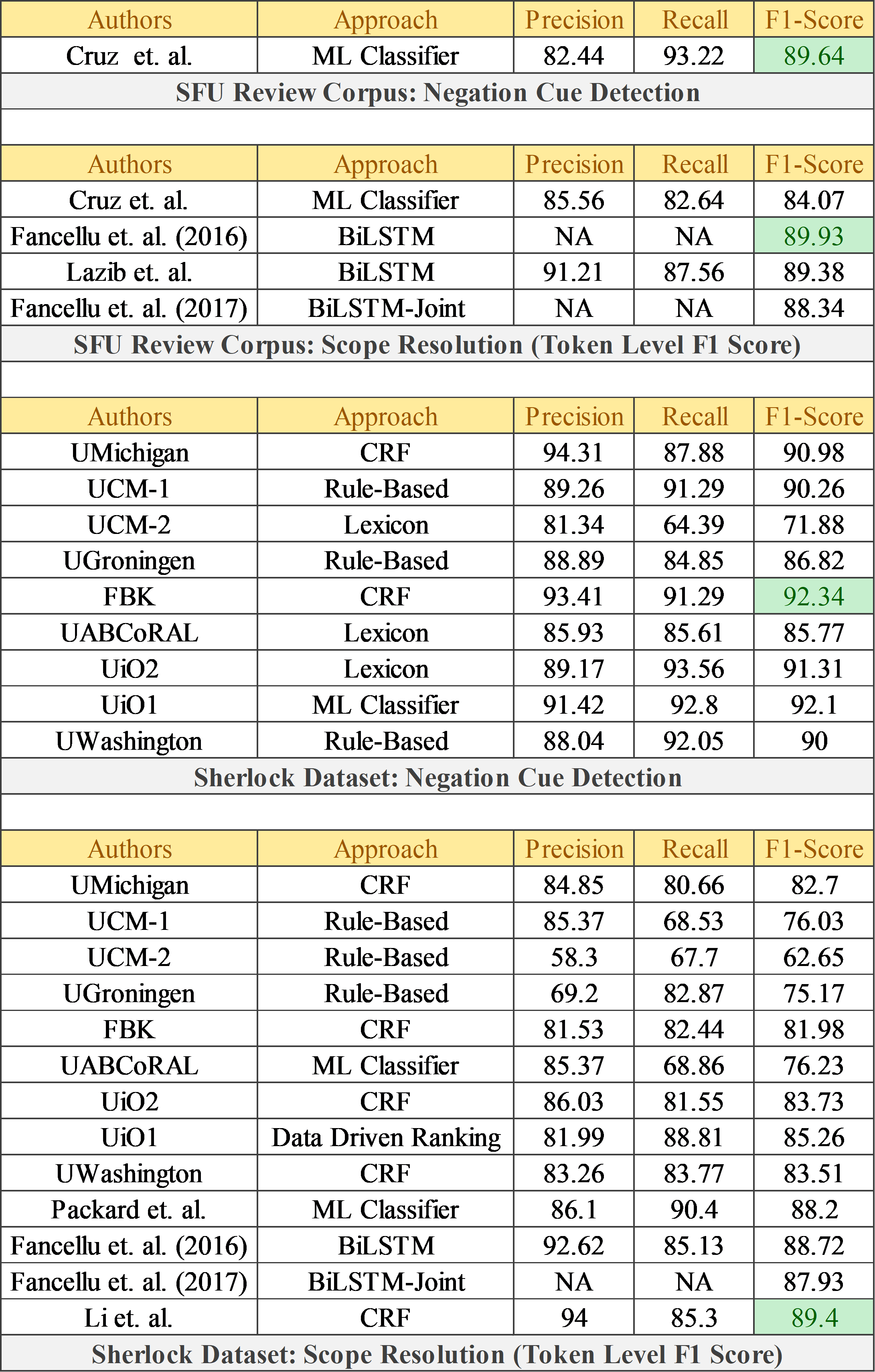}
    \caption{Literature Review: Results Summary}
    \label{fig:litreviewfig2}
\end{table}

\subsection{Deep Learning Approaches}
More recent approaches have looked to apply Deep Learning architectures to the task.  \newcite{qian-etal-2016-speculation} were the first to apply deep learning to negation scope detection. They used Convolutional Neural Networks to path features to generate embeddings, which they concatenated with position features and fed to a softmax layer to compute confidence scores of its location labels. They used this system on the BioScope Corpus and outperformed all existing systems on the BioScope Abstracts.
\par \newcite{fancellu-etal-2016-neural} looked at neural networks for scope detection. They rightly point out that most systems were highly engineered and only tested on the same genre they were trained on. They experimented with a one-hidden layer feed forward neural network and a bidirectional LSTM (BiLSTM) on the Sherlock Dataset, and found that the BiLSTM performed the best.
\par \newcite{lazib} at around the same time looked to Recurrent Neural Network variants for scope resolution. They experimented with RNN, LSTM, BiLSTM, GRU and CRF on the SFU Review Corpus Dataset. The BiLSTM, again, gave the best performance. Thus, we see different datasets benefitting from the use of BiLSTMs, indicating the potential in DL-based methods.
\par \newcite{fancellu-etal-2017-detecting} performed an analysis of the available datasets and showed that there existed a problem which enabled systems to gain high accuracy, namely that negation scopes were frequently annotated as a single span of text delimited by punctuation. They pointed out that the Bioscope and SFU Review corpus suffer from this problem, while the Sherlock Corpus does not. They also improved upon their previous model \cite{fancellu-etal-2016-neural} (BiLSTM) by making joint predictions for all words. Their earlier approach would model the prediction of scope as independent predictions for each word. They added a dependence on the previous prediction for the next. By doing so, they managed to improve the best system for the task.
\par \newcite{fancellu2018} showed that BiLSTMs were the state of the art, and that models suffer from genre(domain) effects. They also looked at cross-lingual scope detection, finding negation scope in languages where annotations aren’t available, which is a common problem for low-resource languages.
\par \newcite{Gautam2018LongST} looked at handling negation in tutorial dialogues. They too looked at LSTMs to solve a sequence labelling problem and got promising results on a custom dataset. Both cue detection and scope resolution were done using LSTMs.
\par \newcite{Taylor2018TheRO} used a combined BiLSTM to label cue and scope simultaneously. They wanted to look to augment patient cohort identification from electroencephalography reports. They preprocessed the text first, and then used the Gensim implementation of word2vec to generate embeddings for the text. Word embeddings were the first attempt at using Transfer Learning in NLP.
\par More recently, \newcite{Bhatia2020} used a shared encoder and 2 separate decoders to get the entities and negations respectively. They performed evaluation over the i2b2/VA dataset and a proprietary medical condition dataset and showed that the joint model outperforms all standard models. They used a BiLSTM to encode the sequence at the word level, and an LSTM decoder. This method showed the power of using a joint encoding for both tasks.
\par \newcite{chen} used attention based BiLSTM networks and word embeddings to detect assertions and negations. This method applied attention, one of the more promising components of architectures addressing other NLP problems, to scope resolution.

\section{Methodology}

\begin{figure}[h]
    \centering
    \includegraphics[width = 0.95\linewidth]{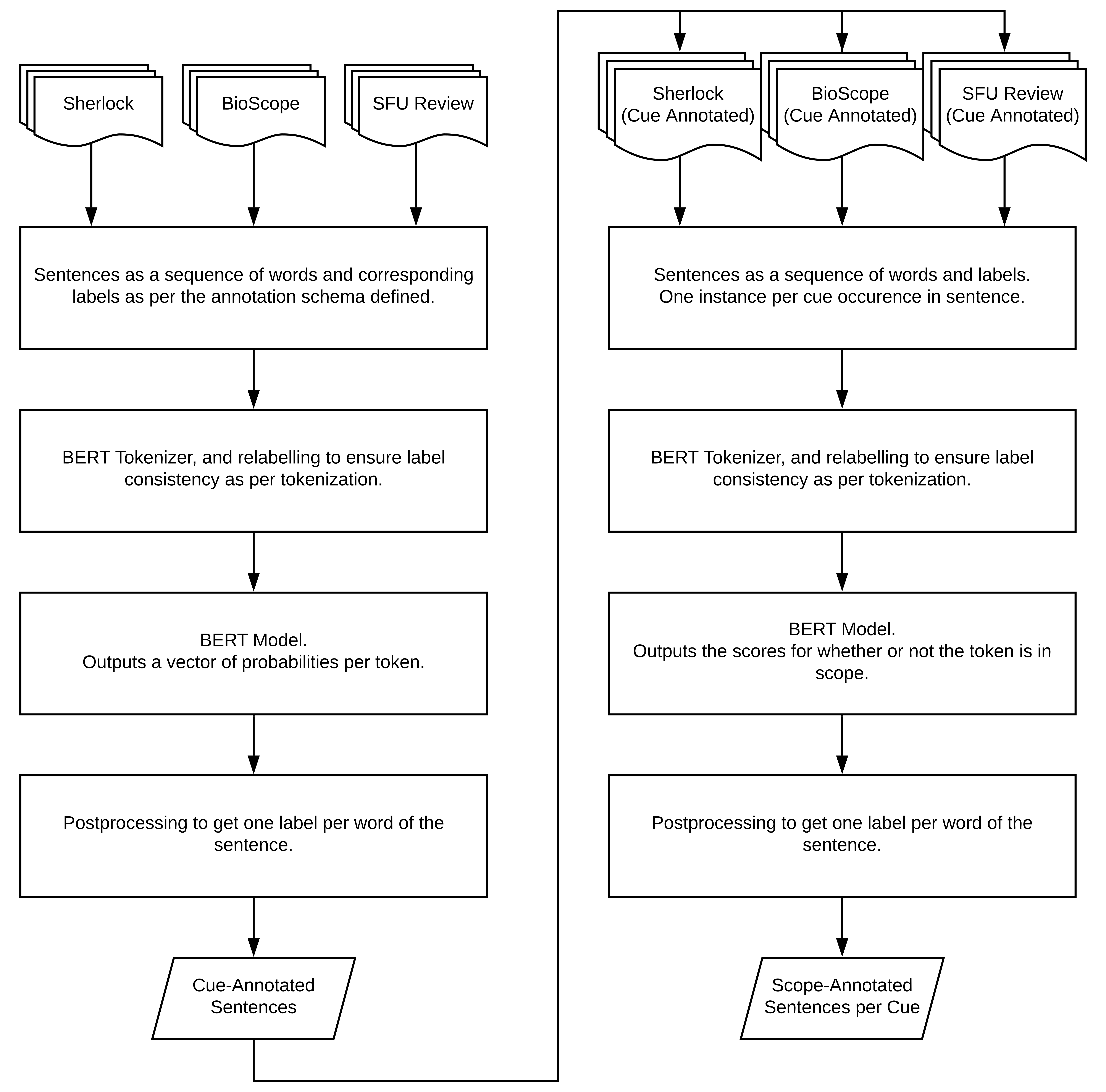}
    \caption{Proposed System Design (NegBERT)}
    \label{fig:negbert}
\end{figure}

We approach the task in the typical 2-stage fashion: negation cue detection performed before scope resolution, as displayed in Figure \ref{fig:negbert}, which depicts a sample flow through the entire system. For both stages, we use Google’s Bidirectional Encoder Representation for Transformers \cite{bert} (BERT-base) with a classification layer on top of it. We use huggingface’s PyTorch implementation of BERT \cite{Wolf2019HuggingFacesTS}, and finetune the bert-base uncased model (110 million parameters) to the training sets.
\par For negation detection, we use the following annotation schema:
\begin{itemize}[noitemsep]
    \item 0 - Affix
    \item 1 - Normal Cue
    \item 2 - Part of a multiword cue
    \item 3 - Not a Cue
\end{itemize}
\par This scheme is useful for the Sherlock dataset which has annotations for affixes, but the BioScope Corpus and SFU Review Dataset do not have annotations for affixes. Hence, when we test inter-dataset performance, we consider cues that are affixes as normal cues, and predictions of affixes as predictions of normal cues. (i.e. 0 and 1 are considered as the same label for the purpose of evaluation). We also use a 5th label for the padded tokens and set the class weights for that token category to 0 to avoid training on it. 
\par For scope resolution, we use a binary labelling scheme, 0 as not a token and 1 as a token. We feed sentences which we know have cues to the model, and to encode that information into the input, we consider 2 methods:
\begin{enumerate}[noitemsep]
    \item \textbf{Replace:} We replace the token which is the cue with another special token which represents the kind of token it is according the cue detection labelling scheme. Thus, ‘[im]polite’ becomes ‘token[0]’, ‘not’ becomes ‘token[1]’, and	‘neither’ and ‘nor’ both become ‘token[2]’.
    \item \textbf{Augment:} We keep the original word and add the special token according to the scheme above immediately before the word. Thus, ‘[im]polite’ becomes ‘token[0] impolite’, ‘not’ becomes ‘token[1] not’, and ‘neither’ becomes ‘token[2] neither’.
\end{enumerate}
We need to preprocess the input to NegBERT, as the tokenization performed by BERT’s BytePairEncoding (BPE) creates a labelling issue. For instance:
\begin{center}
    \begin{verbatim}
        I am not impolite. ->
        I, am, not, im, ##polite.
    \end{verbatim}
\end{center}
\par While we only have 4 labels corresponding to each word for the sentence, BERT has to be fed 5 labels (1 label per token it is fed). Thus, the token labels are the same as the word from which they were originally split, i.e. a single word cue labelled as ‘1’ can get split into multiple tokens, each having the label ‘1’ denoting a single word cue.
\par Postprocessing is also needed for converting the token-level predictions to word-level predictions, as the output of this system needs to be interpretable. The tokenization via BPE will not make much sense to a layperson using the system. Thus, our postprocessing enables the use of further processing techniques on a sentence that has been annotated with negation cues and their respective scopes.
\par We consider the output for each token as a probability distribution over the classes possible and average them out for all tokens in a word, giving us a probability distribution for a word over all classes of tokens. A simple argmax gives us our required token type.
\par An example flow of a sentence through the proposed system is as follows:
\begin{center}
    \textbf{Input sentence:}  This is not a negation\\
    \textbf{Cue Detection Labels:} [3,3,1,3,3] \\
    After preprocessing (BERT Tokenizer): \\
    \textbf{Input:}[Th, \#\#is, is, not, a, nega, \#\#tion, $\langle pad \rangle, \langle pad\rangle, \langle pad\rangle,...]$ \\ (padding needed to make the number of tokens equal to the number of input tokens to BERT.)\\
    \textbf{Target Output (Labels):} [3,3,3,1,3,3,3,4,4,4,4, ...]\\
    This is input to the model, and the class weights ensure we do not train on the pad tokens (5th label). \\
    \textbf{NegBERT Output: }[[0.01,0.01,0.01,0.95,0.02], [0.01,0.01,0.01,0.95,0.02], [0.01,0.01,0.01,0.95,0.02], [0.02,0.9,0.02,0.02,0.02], [0.01,0.01,0.01,0.95,0.02], [0.01,0.01,0.01,0.95,0.02], [0.01,0.01,0.01,0.95,0.02], x,x,x,....]
\end{center}
\par The x’s are probability vectors for tokens corresponding to padding. Since the target labels corresponding to the pad tokens are set to 4, setting a class weight of 0 will, effectively, not add the loss to the training loss, thus avoiding training on them.
\par We now postprocess this to get the output for the sentence.
\par After the postprocessing,\\
    \textbf{Output:}[3,3,1,3,3]. These correspond to words of the sentence.
\par We perform the preprocessing (split words into tokens, duplicate the labels for the word to the tokens it’s split into) and postprocessing (combining the outputs of multiple tokens that a word was split into to get word-level labels) for scope resolution as well. This postprocessing step increased the F1 score of NegBERT compared to considering just the first label of the first token that a word was split into as its label, indicating that it is important to consider the output for all tokens that a word has been split into.

\section{Experimentation Details}
We use Google’s BERT \cite{bert} (bert-base-uncased) as the base model to generate contextual embeddings for the sentence. The input to the BERT model is a sequence of tokenized and encoded tokens of a sentence. We then use a vector of dimension R\textsuperscript{HxN\_C} (H = hidden state dimensionality of base model, N\_C = number of classes) to compute scores per token, for the classification task at hand. BERT outputs a vector of size R\textsuperscript{H} per token of the input, which we feed to a common classification layer of dimension R\textsuperscript{Hx5} for cue detection and R\textsuperscript{Hx2} for scope resolution. We use early stopping on dev data for 6 epochs as tolerance and F1 score as the early stopping metric, use the Adam optimizer with an initial learning rate of 3e-5, and the Categorical Cross Entropy Loss with class weights as described above to avoid training on the padded label outputs.
\par We perform cue detection and scope resolution for all 3 datasets, and train on 1 and test on all datasets. For the Sherlock dataset, the training data is the Sherlock Train data used in *sem 2012 Shared Task available in cd-sco. The dev data is the dev data provided in the Sherlock Corpus, and the test data is the Sherlock Cardboard and Circle data used as test data for *sem 2012. For all other corpora, we use a default 70 – 15 – 15 split for the train-dev-test data. We trained the models on free GPUs available via Google Colaboratory, the training scripts are publicly available\footnote{https://adityak6798.github.io/}.

\section{Results}
The results are tabulated in Tables \ref{fig:negbertcueresults} and \ref{fig:negbertscoperesults}.

\begin{table}[h]
    \centering
    \includegraphics[width = 0.95\linewidth]{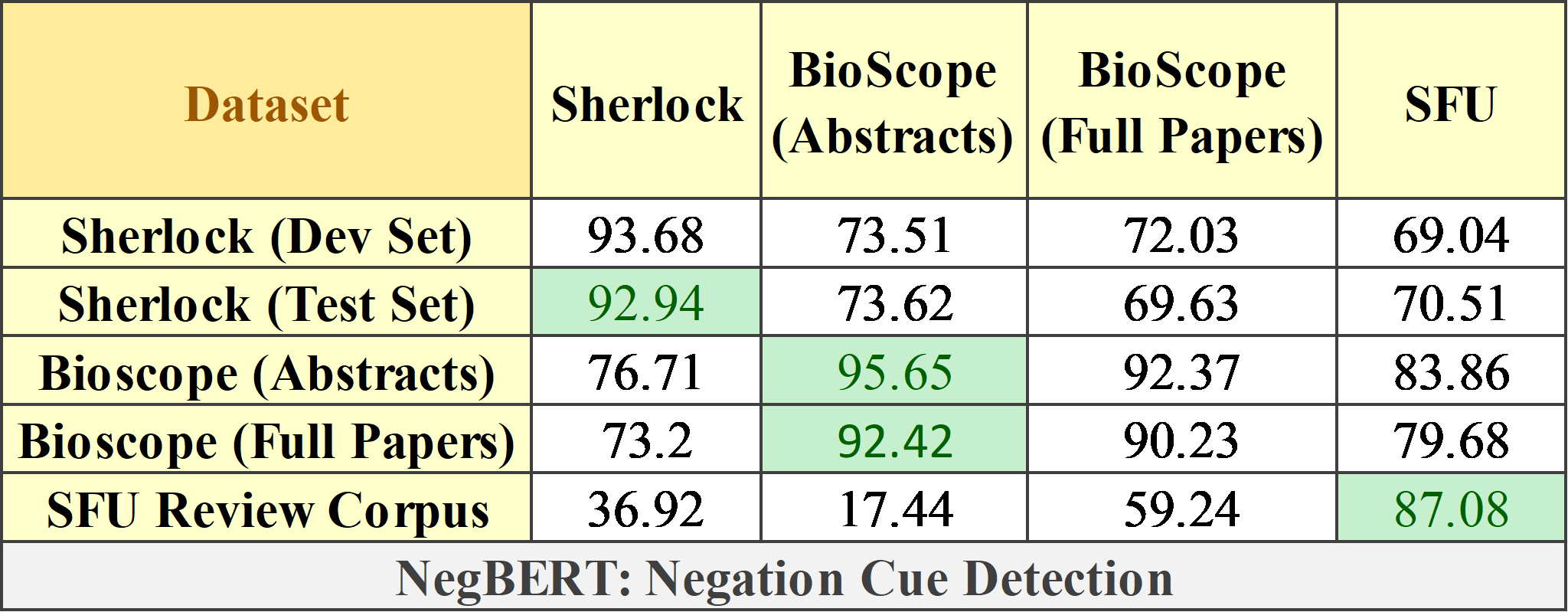}
    \caption{Results for Cue Detection: The row represents the test dataset, the column represents the train dataset}
    \label{fig:negbertcueresults}
\end{table}

\par For cue detection, on the Sherlock dataset test data, we see that we outperform the best system \cite{chowdhury-2012-fbk} by 0.6 F1 measure. On the BioScope Abstracts subcorpus, we perform reasonably well, achieving an F1 of 95.65. On the BioScope Full papers subcorpus, we are able to achieve 90.23 F1 when training on the same data, but we do note that the amount of training data available is significantly lower than for the other datasets, and while general Deep Learning based approaches cannot perform well in such situations, we still manage to perform well. Unlike Morante and Daelemans, we do so without using a word lexicon taken from the data itself, thus allowing the model to generalize, as seen in its performance on BioScope Full Papers subcorpus (F1: 92.42) when trained on the BioScope Abstracts subcorpus. On the SFU Review Corpus, we achieve an F1 of 87.08.
\par For the inter-dataset comparison, we note that the model generalizes well across different domains, except the SFU Review corpus. We think this is due to annotation differences in both datasets, and that SFU corpus has cues that the other corpora do not have.

\begin{table}[h]
    \centering
    \includegraphics[width = 0.95\linewidth]{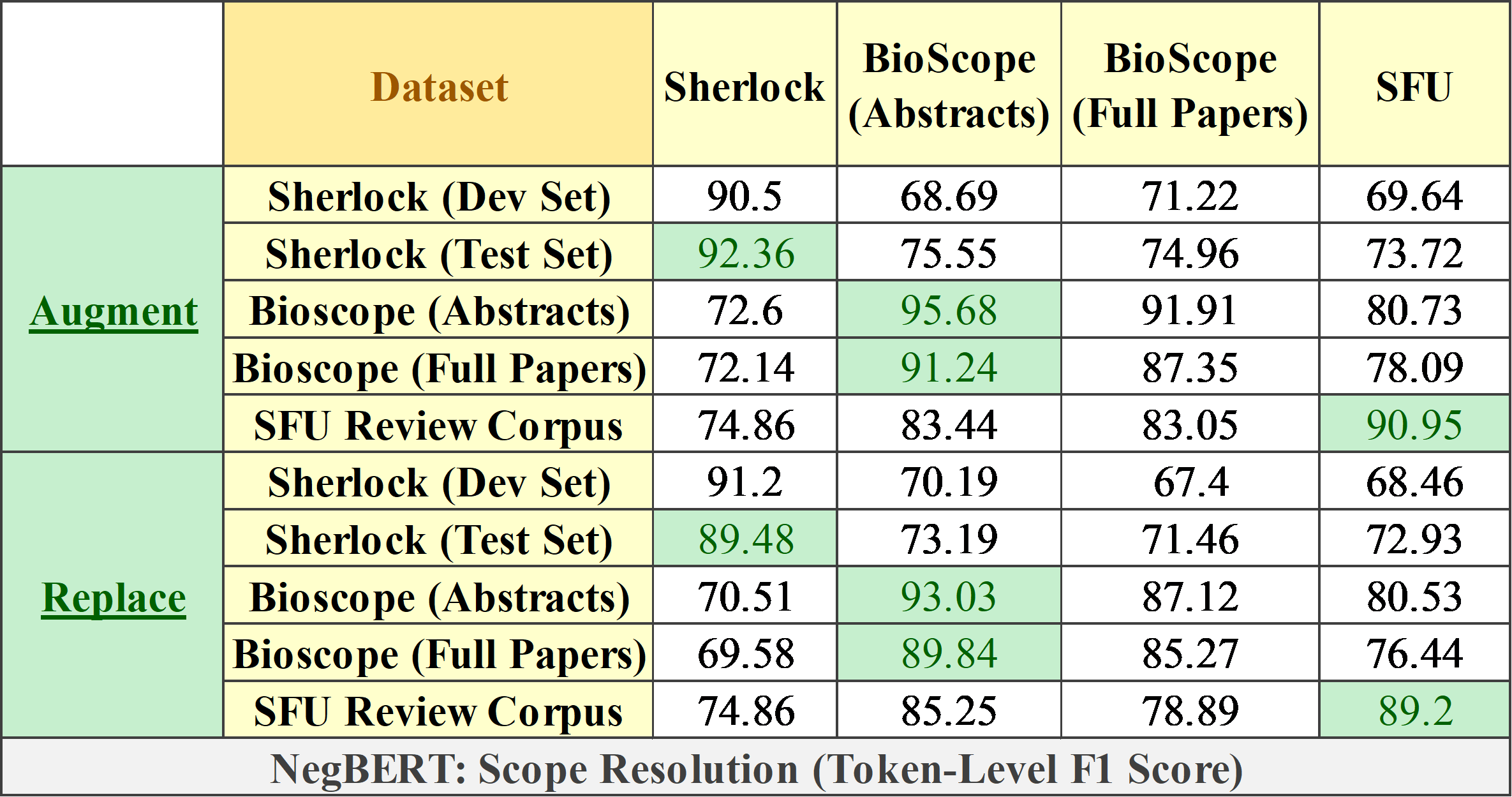}
    \caption{Results for Scope Resolution: The row represents the test dataset, the column represents the train dataset}
    \label{fig:negbertscoperesults}
\end{table}

\par For scope resolution: On the Sherlock dataset, we achieve an F1 of 92.36, outperforming the previous State of the Art by a significant margin (almost 3.0 F1). On the BioScope Abstracts subcorpus, we achieve an F1 of 95.68, outperforming the best architecture by 3.57 F1. On the Bioscope Full Papers subcorpus, we outperform the best architecture by 2.64 F1 when training on the same dataset, and gain an additional performance improvement of 3.89 F1 when trained on the BioScope Abstracts subcorpus. On the SFU Review Corpus, we outperform the best system to date by 1.02 F1.

\section{Analysis}

\begin{table}[h]
    \centering
    \includegraphics[width = 0.95\linewidth]{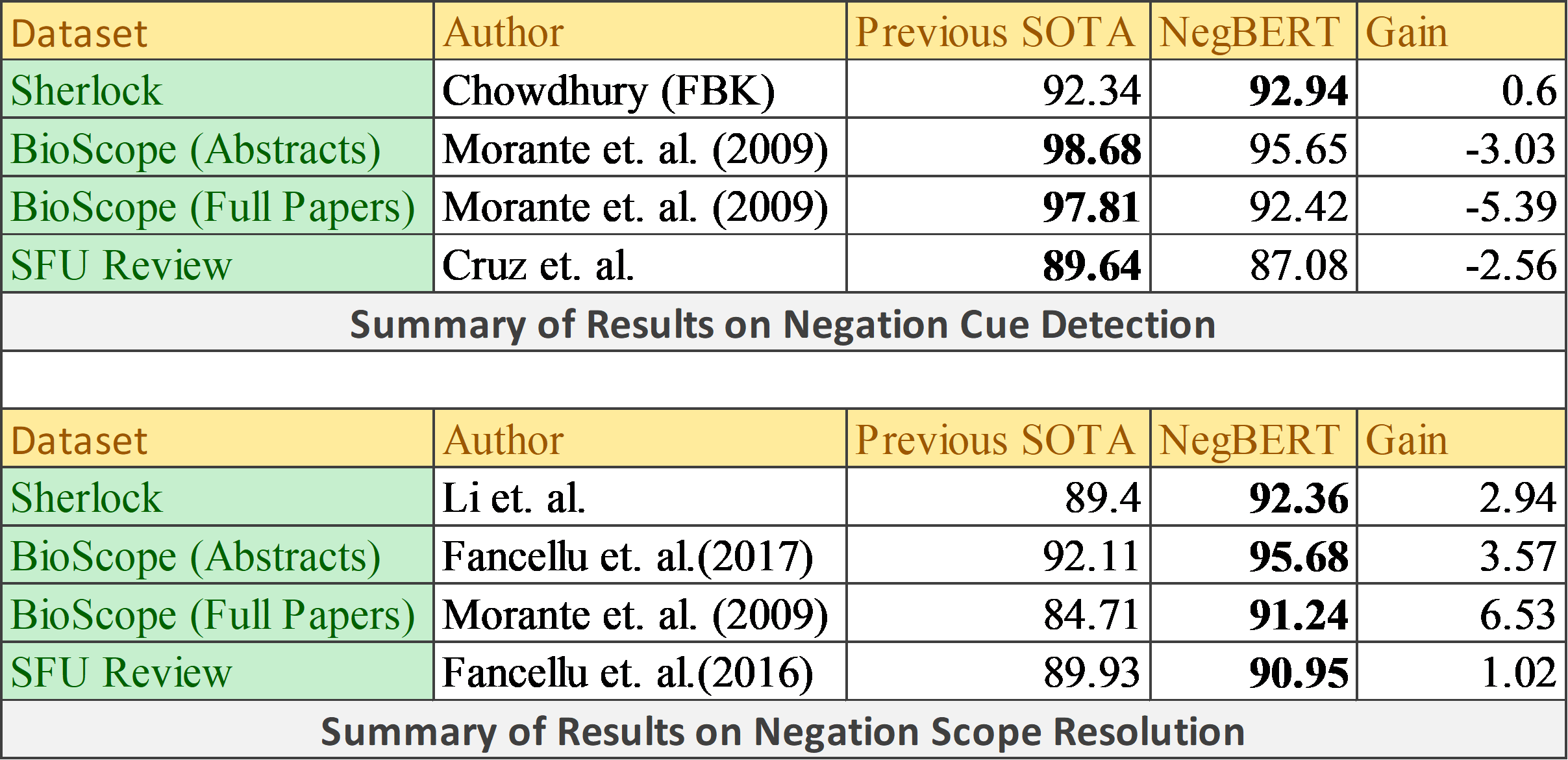}
    \caption{NegBERT Results Comparison with State-of-the-Art results}
    \label{fig:resultscomparison}
\end{table}

The comparison of NegBERT's results with the state-of-the-art systems for negation cue detection and scope resolution is presented in Table \ref{fig:resultscomparison}. For negation cue detection, we observe a significant gap between our model, NegBERT, and the current state-of-the-art systems, while we outperform the baseline systems. We believe this is so as these datasets are fairly limited in size and scope, and for a such a task, bigger models like BERT need a lot more examples to train on to master the finer points of negation detection, while this is straightforward to handle for rule-based approaches and smaller datasets. This model does outperform other Deep Learning based systems applied to negation cue detection.
NegBERT’s gain in accuracy on Scope Resolution is because it allows contextual embeddings and knowledge transfer across millions of documents to downstream tasks.
\par When we trained on BioScope Abstracts and tested on the BioScope Full Papers, we surprisingly observed a state-of-the-art result of 91.24 (a gain of 3.89 F1 points over training on BioScope Full Papers), which is far beyond the achievable results on training and evaluating on the BioMedical sub corpora. This is only possible because of BERT’s pretraining, and the similarity of the sub corpora of the BioScope Corpus. 
\par We also notice that in general, though the cross-dataset generalizability is acceptable, it is far from what one would desire. We believe that  the combination of these 2 results  indicate that the datasets are highly disjoint in their representations of negations and are fairly limited in size, both of which contribute to the system’s inability to perform well on unseen data from a different domain, but perform well on data from within the same domain.

\section{Error Analysis}

\newcite{fancellu-etal-2017-detecting} suggest that most systems trained until now on the BioScope Corpus and the SFU Review Corpus suffer from learning surface patterns based on punctuations, as scopes of sentences are delimited by punctuations for a high percentage of the sentences in these two datasets. This bad annotation of these 2 corpora makes most systems vulnerable to this pattern.
\par We tested NegBERT for this phenomenon by comparing the results for sentences whose scope is delimited by punctuation versus sentences whose scope isn’t by creating 2 sub-corpora per dataset: one where the scope is delimited by punctuation and the other where it isn’t. We find out if a sentence has its scope delimited by punctuation by the following process:
\begin{itemize}[noitemsep]
    \item Split the sentence into words by splitting on a space.
    \item Find the index of the last token before the first cue occurrence in the sentence containing a punctuation (let it be P\_first), and the index of the first token after the last cue occurrence containing a punctuation (let it be P\_last).
    \item If the P\_first is equal to the index of the first token in scope or 1 less than that, or if P\_last is equal to the index of the last token in scope or is 1 more than that, this sentence is said to belong to the punctuation subcorpus, where the scope is delimited by punctuation (i.e. The scope is helped by the presence of punctuation.)
    \item Symbol list: \\
    \verb1!"#$%&\'()*+,-./:;<=>?@[\\]^_`{|}~1

\end{itemize}
\par We look at the Percentage of correct scopes (PCS), i.e. the percentage of sentences where there is an exact scope match. Mislabeling as little as 1 token of a sentence is considered as wrong scope annotations. The results are shown in Table \ref{fig:negberterroranalysis}.

\begin{table}[h]
    \centering
    \includegraphics[width = 0.75\linewidth]{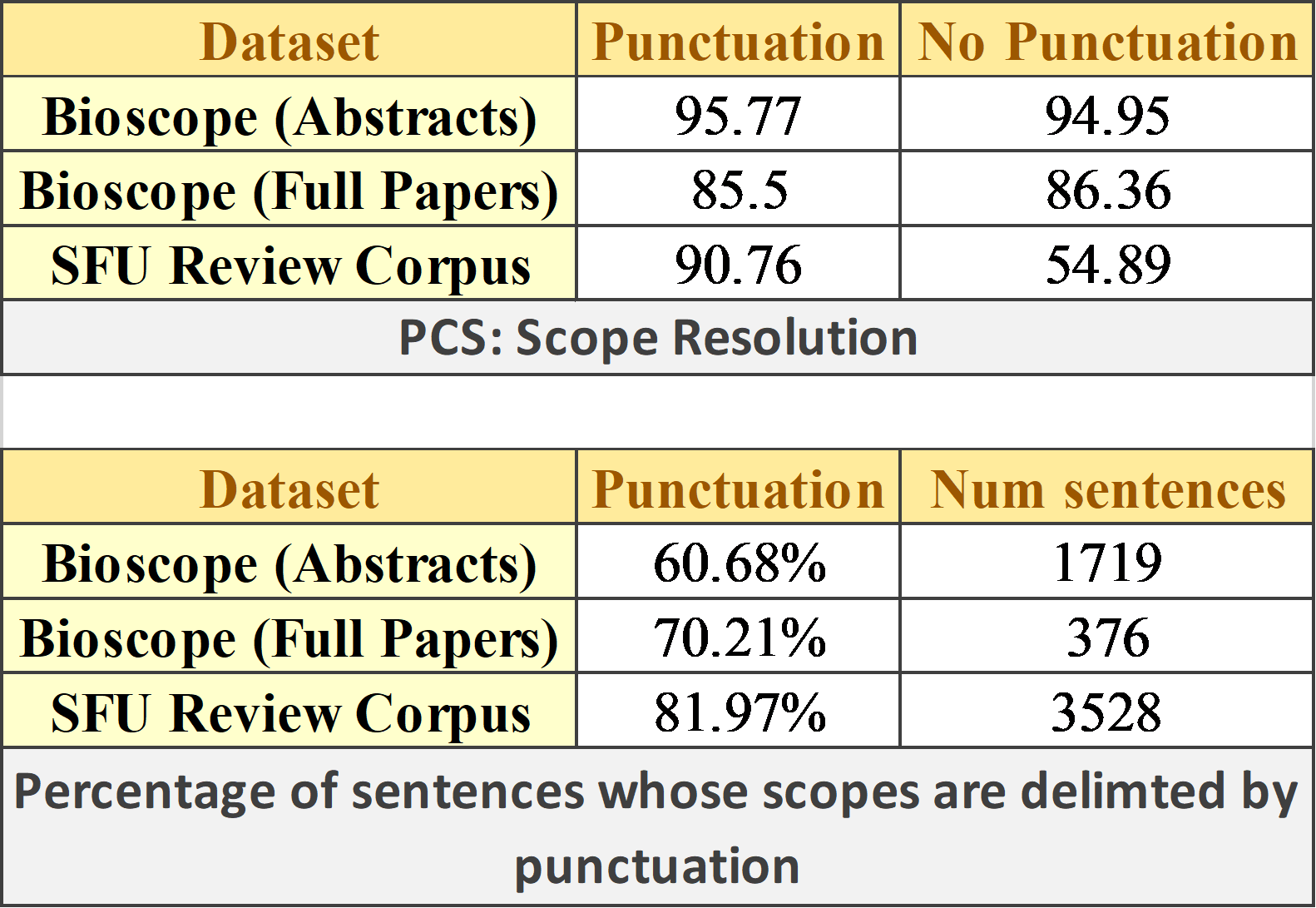}
    \caption{Results for Error Analysis}
    \label{fig:negberterroranalysis}
\end{table}

\par We observe that BERT performs almost similarly on the punctuation and no punctuation corpora for the BioScope Corpora, while the SFU corpora does suffer from this problem due to a significantly lower percentage of training instances of the no punctuation kind.
\par When NegBERT was trained solely on the sentences from the SFU Corpus whose scope are not delimited by punctuation (70-15-15 split of this sub-corpus, same procedure used to train), no gain in PCS was observed (PCS: 54.16), indicating that the reason for the poor performance on this dataset is because of insufficient data, and not because the model overfits to a more frequently occurring phenomena. 
\par NegBERT does not suffer from the poor annotation problem when trained on the BioScope Corpus and achieves the best reported result on the Sherlock corpus, which is does not suffer from this poor annotation problem.
\par Thus, we conclude that NegBERT indeed learns more than surface patterns of scopes based on punctuations on the BioScope Corpus and the Sherlock Corpus. We attribute the poor performance on the non-punctuation delimited SFU Review Corpus to an insufficient number of such sentences available for training from the SFU Review Corpus.

\section{Conclusion and Future Scope}
\par Negation Cue Detection and Scope Resolution is a very well researched problem. We reviewed all existing papers and identified the research trends moving towards Deep Learning approaches. Following the general trend in the NLP community, we looked to the new generation of transfer learning models (BERT) to solve both tasks. We explored the set of design choices and reported a significant improvement in scope detection systems using BERT-base uncased model. We also analyzed the inter-domain generalization of the models, and noted that the use of our proposed architecture, NegBERT, as the underlying model allows for really good performance on scope resolution for unseen datasets from different domains. We reported a new state-of-the-art model on every publicly available dataset using the same architecture with no task-specific tuning and the same set of hyperparameters for scope resolution. 
\par We then performed an error-analysis on the BioScope and SFU Review Corpora, which suffer from poor annotation leading to most models picking up surface patterns \cite{fancellu-etal-2017-detecting}, but report that NegBERT only suffers from this problem on the SFU Review Corpus to a certain degree, and avoids this problem on the BioScope Corpus, learning more than just surface patterns. Thus, we clearly establish the usefulness of pretrained models and the usage of transfer learning to the task of negation scope resolution.
\par We envision that the future progress in this task should focus on the use of ever-changing state-of-the-art models in the transfer learning domain which have significant potential to improve the accuracy of the system. We feel that a bigger dataset is needed to extract the maximum generalizability from such architectures.
\section{Bibliographical References}\label{reference}

\bibliographystyle{lrec}
\bibliography{lrec2020W-xample-kc}

\end{document}